\documentclass[a4paper, 10pt, conference]{ieeeconf}      
\usepackage{FG2024}
\usepackage{subcaption}
\usepackage{graphicx}
\usepackage{booktabs}
\usepackage{textcomp}
\FGfinalcopy 

\IEEEoverridecommandlockouts                              
\usepackage{graphics} 
\usepackage{epsfig} 
\usepackage{mathptmx} 
\usepackage{times} 
\usepackage{amsmath} 
\usepackage{amssymb}  

\title{\LARGE \bf
1st Place Solution to the 1st SkatingVerse Challenge
}

\author{\parbox{16cm}{\centering
   {Tao Sun\textsuperscript{*}, Yuanzi Fu\textsuperscript{*}, Kaicheng Yang\textsuperscript{*}, Jian Wu\textsuperscript{*}, Ziyong Feng \\
   DeepGlint \\
   {\tt\small \{taosun,yuanzifu,kaichengyang,jianwu,ziyongfeng\}@deepglint.com}} \qquad
   }
}

\begin{document}

\ifFGfinal
\thispagestyle{empty}
\pagestyle{empty}

\pagestyle{plain}
\fi

\maketitle

\renewcommand{\thefootnote}{\fnsymbol{footnote}}
\footnotetext[1]{These authors contributed equally in this work.}

\begin{abstract}

This paper presents the winning solution for the 1st SkatingVerse Challenge. We propose a method that involves several steps. To begin, we leverage the DINO framework to extract the Region of Interest (ROI) and perform precise cropping of the raw video footage.  Subsequently, we employ three distinct models, namely Unmasked Teacher, UniformerV2, and InfoGCN, to capture different aspects of the data. By ensembling the prediction results based on logits, our solution attains an impressive leaderboard score of 95.73\%.

\end{abstract}

\section{INTRODUCTION}

\begin{figure}[h]
\includegraphics[width=0.9\linewidth]{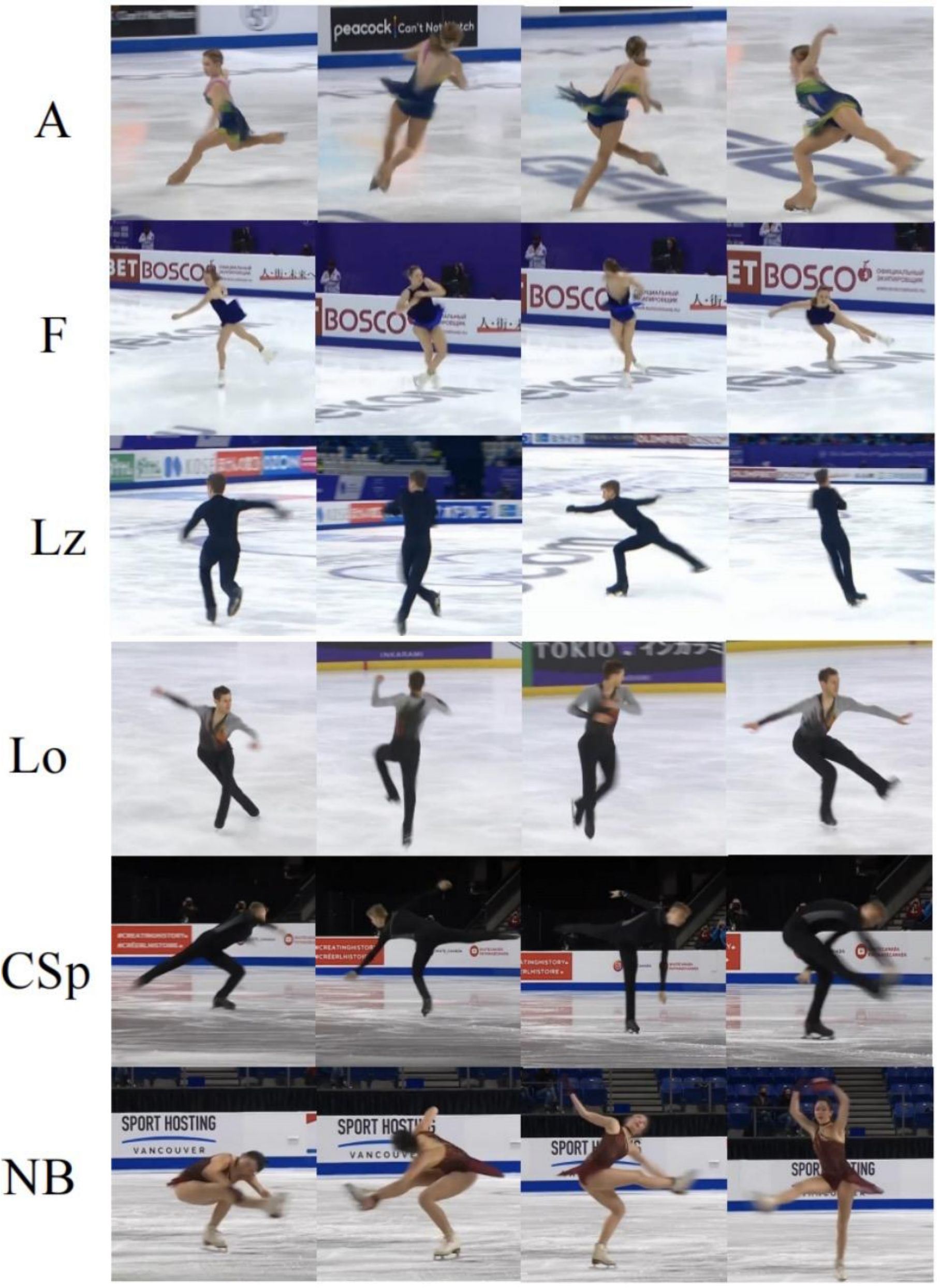}
\caption{Examples of figure skating actions: A (Axel); Lo (Loop); F (Flip); CSp (Camel Spin); Lz (Lutz); NB (No Basic)\protect\footnotemark[2].}
\label{fig1}
\end{figure}

The 1st SkatingVerse Challenge, held as part of the SkatingVerse Workshop, is affiliated with the prestigious 18th IEEE International Conference on Automatic Face and Gesture Recognition (FG). In contrast to previous tasks that may lack practical applicability, such as fine-grained action segmentation and assessment, this challenge focuses on the construction of a comprehensive dataset comprising 1,687 continuous videos from figure skating competitions. The primary objective is to foster the development of algorithms capable of accurately analyzing each action depicted in these videos. The challenge dataset consists of 19,993 video clips for training and 8,586 video clips for testing. It encompasses a wide range of 28 distinct categories of figure skating actions, and Figure~\ref{fig1} illustrates six representative examples of these actions.

\renewcommand{\thefootnote}{\fnsymbol{footnote}}
\footnotetext[2]{Examples are from the challenge website:https://skatingverse.github.io/}

\section{METHOD}
\subsection{Dataset Pre-processing}
In order to enhance the model's attention toward figure skating actions, we performed human body detection on the original video frames. To accomplish this, we utilized FFmpeg to extract video frames and employed the DINO framework~\cite{zhang2022dino} for extracting bounding boxes corresponding to human detections in each frame. These individual bounding box results from all frames within a video were then consolidated to generate the ultimate detection box for that specific video. Finally, using FFmpeg, we crop the raw video clips based on the combined bounding box information obtained from the human detection process.

\subsection{Model Structure}
\begin{figure*}
\includegraphics[width=0.95\linewidth]{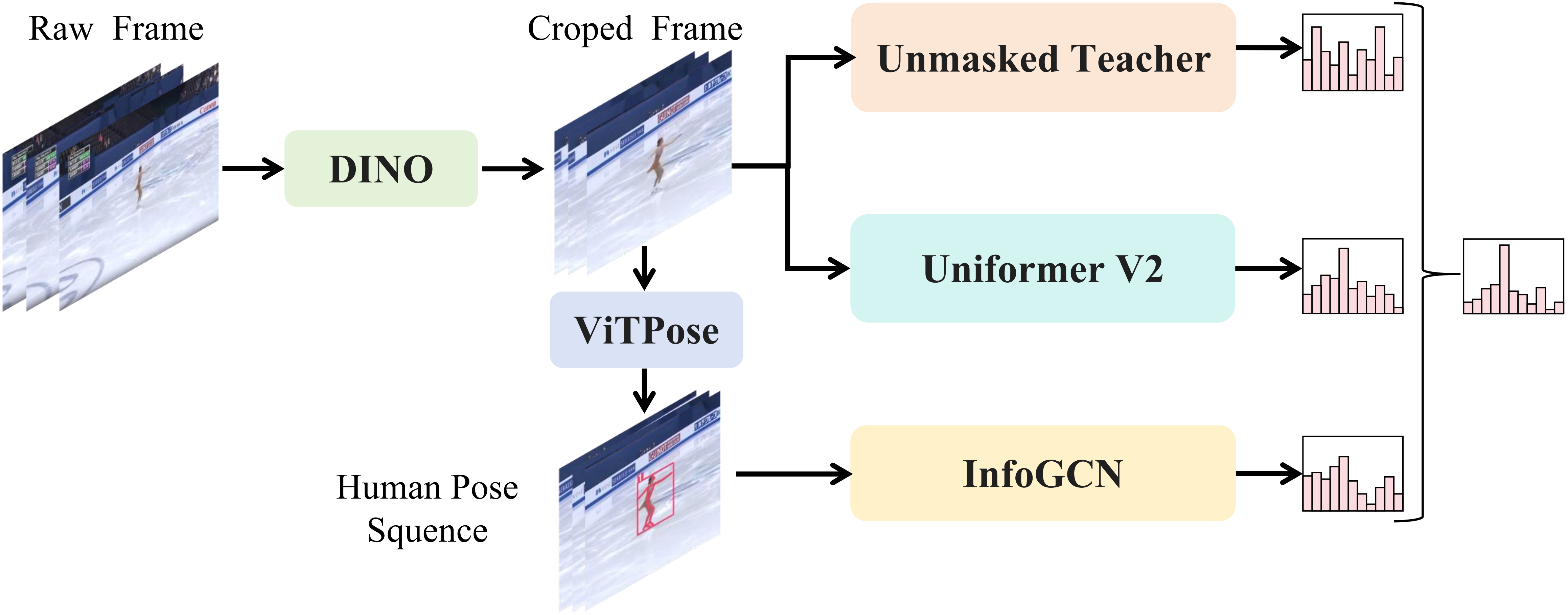}
\caption{The overall architecture of our solution.}
\label{fig2}
\end{figure*}

The architecture of our proposed method is illustrated in Figure~\ref{fig2}. To begin, we employ the DINO model~\cite{zhang2022dino} to extract precise human detection bounding boxes, facilitating the generation of cropped frames. Subsequently, we conduct fine-tuning on two widely-used general video pre-training models, namely Unmasked Teacher~\cite{li2023unmasked} and UniformerV2~\cite{li2022uniformerv2}. Furthermore, we leverage ViTPose~\cite{xu2022vitpose} to extract human skeleton sequences, which are then utilized to train the InfoGCN model~\cite{chi2022infogcn} for accurate action predictions.

\subsubsection{Unmasked Teacher}
Unmasked Teacher(UMT)~\cite{li2023unmasked} is a two-stage training-efficient pretraining framework that enhances temporal-sensitive video foundation models by incorporating the advantages of previous approaches. In Stage 1, the UMT exclusively utilizes video data for masked video modeling, resulting in a model that excels at video-only tasks. In Stage 2, the UMT leverages public vision-language data for multi-modality learning, enabling the model to handle complex video-language tasks such as video-text retrieval and video question answering. 

In this paper, we initially fine-tune the pre-trained UMT-L16 model (which is pre-trained and fine-tuned on Kinetices710 with $8 \times 224^2$ input images) for 50 epochs using $16 \times 224^2$ input images. Subsequently, based on the obtained fine-tuned model weights, we perform weight interpolation and further fine-tune the model for 10 additional epochs. This fine-tuning is conducted separately using both $16 \times 448^2$ images and $32 \times 224^2$ images. To obtain the final prediction result of UMT, we aggregate the predicted probabilities from the three models and apply the softmax function.

\subsubsection{UniformerV2}
UniformerV2~\cite{li2022uniformerv2} is a versatile approach for building a robust collection of video networks. It combines the image-pretrained Vision Transformers (ViTs) with efficient video designs from UniFormer~\cite{li2022uniformer}. The key innovation in UniformerV2 lies in the inclusion of novel local and global relation aggregators. These aggregators seamlessly integrate the strengths of both ViTs and UniFormer, achieving a desirable balance between accuracy and computational efficiency.

In this work, we conduct fine-tuning on the UniFormerV2-L14 model(we use the CLIP~\cite{radford2021learning} model weight pre-trained on LAION400M~\cite{schuhmann2021laion}) for 30 epochs, utilizing $32 \times 224^2$ input images.

\subsubsection{ViTPose \& InfoGCN}
To obtain additional skating-related information from human skeleton sequences, we first use ViTPose~\cite{xu2022vitpose} to extract the human skeleton sequence for each cropped frame. ViTPose utilizes plain and non-hierarchical vision transformers as backbones to extract features for individual person instances. It also incorporates a lightweight decoder for efficient pose estimation. ViTPose offers great flexibility in terms of attention type, input resolution, pre-training, fine-tuning strategies, and the ability to handle multiple pose tasks.

Following the extraction of the human skeleton sequence for each cropped frame using ViTPose, we obtain a sequence of length 320 for each video. To accurately predict the categories of figure skating actions, we train the InfoGCN~\cite{chi2022infogcn} model from scratch over the course of 300 epochs. This training process enables the model to effectively analyze and classify the various figure skating actions depicted in the videos.

\section{Experiments}
\subsection{Implementation Details}
All experiments in this study are conducted using the PyTorch 2.0 framework, and the training process is performed on a powerful setup consisting of 8 NVIDIA A100 (80G) GPUs. The input image size is set to 224 × 224 pixels. To optimize the training process, we utilize the AdamW optimizer~\cite{loshchilov2017decoupled}. The specific hyperparameters employed for fine-tuning UMT, UniformerV2, and InfoGCN are provided in Table~\ref{umt_hyperparam}, offering detailed insights into the configuration choices.

To expedite the training process, we incorporate FlashAttention2.0~\cite{dao2023flashattention2}, which leads to a notable 2 to 3 times acceleration in training speed. Additionally, we employ gradient accumulation, further enhancing the efficiency of the training process. During model inference, we adopt Test-Time Augmentation (TTA), which involves conducting 5 temporal and 3 spatial samplings, resulting in a total of 15 predictions. These predictions are subsequently fused together to generate the final prediction result, ensuring robustness and accuracy in the model's output.

\begin{table*}[h]
\centering
\caption{Detail hyperparameters used in training UMT, UniformerV2, and InfoGCN.}
\subfloat[
\small{Unmasked Teacher-L16 with $16 \times 224^2$ input images.}
]{
\centering
\begin{minipage}{0.3\linewidth}{\begin{center}
\resizebox{\textwidth}{!}{
\begin{tabular}{l|c}
\toprule Hyperparameter & Value \\
\midrule
Adam $\beta_{1}$ & $0.9$ \\
Adam $\beta_{2}$ & $0.98$ \\
Adam $\epsilon$  & $10^{-6}$ \\
Weight decay & $0.05$ \\
Drop path & 0.2 \\
Layer decay & 0.85 \\
Batch size & 128 \\
Learning rate & 2e-4 \\
Warmup epochs & 5 \\
Training epochs & 50 \\
Gradient accumulation & 2 \\
GPU & $8 \times $A100(80G) \\
\bottomrule
\end{tabular}}
\end{center}}\end{minipage}}
\hspace{2mm}
\subfloat[\small{UniformerV2-L14 with $32 \times 224^2$ input images.}
]{
\centering
\begin{minipage}{0.3\linewidth}{\begin{center}
\resizebox{\textwidth}{!}{
\begin{tabular}{l|c}
\toprule Hyperparameter & Value \\
\midrule
Adam $\beta_{1}$ & $0.9$ \\
Adam $\beta_{2}$ & $0.999$ \\
Adam $\epsilon$  & $10^{-8}$ \\
Weight decay & $0.05$ \\
Drop path & 0.2 \\
Batch size & 16 \\
Learning rate & 2e-5 \\
Warmup epochs & 5 \\
Training epochs & 30 \\
GPU & $4 \times $A100(80G) \\
\bottomrule
\end{tabular}}
\end{center}}\end{minipage}}
\hspace{2mm}
\subfloat[\small{InfoGCN with input human skeleton sequences.}
]{
\centering
\begin{minipage}{0.3\linewidth}{\begin{center}
\resizebox{\textwidth}{!}{
\begin{tabular}{l|c}
\toprule Hyperparameter & Value \\
\midrule
Adam $\beta_{1}$ & $0.9$ \\
Adam $\beta_{2}$ & $0.999$ \\
Adam $\epsilon$  & $10^{-8}$ \\
Weight decay & $0.0005$ \\
Batch size & 128 \\
Learning rate & 1e-3 \\
Warmup epochs & 5 \\
Training epochs & 300 \\
GPU & $1 \times $A100(80G) \\
\bottomrule
\end{tabular}}
\end{center}}\end{minipage}}
\label{umt_hyperparam}
\end{table*}

\subsection{Evaluation Metric}
In this challenge, the performance of the model is assessed based on the mean accuracy of the categories, denoted as $Mean$. The number of correctly predicted samples, referred to as $M$, represents the instances where the category with the highest confidence from the model aligns with the actual category. The total number of samples is denoted as $N$. For each category with $N_i$ samples, if the number of correctly predicted samples is $M_i$, the mean accuracy $Mean$ is calculated using the following formula:
\begin{equation}
\text {$Mean$}=\frac{1}{l} \sum_{i=1}^l \frac{M_i}{N_i}
\end{equation}

\subsection{Experiment Result}
Tab.~\ref{result1} showcases the experimental results on the leaderboard, highlighting the performance of three distinct models. Notably, both Unmasked Teacher (UMT) and UniformerV2 demonstrate a considerable improvement in performance compared to InfoGCN. This notable enhancement can be attributed to the fact that UMT and UniformerV2 are video foundation models, benefiting from pre-training on extensive video datasets.
\begin{table}[h]
\caption{Summary of experimental results on the leaderboard.}
\vspace{1mm}
\label{result1}
\centering
    \resizebox{0.7\columnwidth}{!}{
    \begin{tabular}{lcccc}
    \toprule
    Method  & Online Score  \\
    \midrule
    Unmasked Teacher~\cite{li2023unmasked} & 94.55 \\
    UniformerV2~\cite{li2022uniformerv2} & 95.02 \\
    InfoGCN & 92.03 \\
     \bottomrule
    \end{tabular}}
\vspace{-4mm}
\end{table}

\subsection{Model Ensemble}
To further improve the leaderboard score, we adopt two simple ensemble strategies. The first approach involves voting among all the model predictions, with UniformerV2's prediction being selected in case of a tie. The second approach involves weighted aggregation of the predictions, where the weights for UMT, UniformerV2, and InfoGCN are determined using the softmax function with weights [94.5, 95.0, 92.0], respectively. The ensemble results are shown in Tab.~\ref{result2}

\begin{table}[h]
\caption{Summary of ensemble results on the leaderboard.}
\vspace{1mm}
\label{result2}
\centering
    \resizebox{0.7\columnwidth}{!}{
    \begin{tabular}{lcccc}
    \toprule
    Method  & Online Score  \\
    \midrule
    Model Voter & 95.67 \\
    Model Weighted Summation & 95.73 \\
     \bottomrule
    \end{tabular}}
\vspace{-4mm}
\end{table}

{\small
\bibliographystyle{ieee}
\bibliography{sample_FG2024}
}

\end{document}